\def\year{2020}\relax
\documentclass[letterpaper]{article} 
\usepackage{aaai20}  
\usepackage{times}  
\usepackage{helvet} 
\usepackage{courier}  
\usepackage[hyphens]{url}  
\usepackage{graphicx} 
\usepackage{graphicx}  
\usepackage{mathtools}
\DeclarePairedDelimiter{\floor}{\lfloor}{\rfloor}
\usepackage{multirow}
\usepackage{xcolor}
\usepackage{kotex}
\usepackage{booktabs}
\usepackage{amssymb}

\newcommand{\ie}{\textit{i}.\textit{e}.}
\newcommand{\eg}{\textit{e}.\textit{g}.}

\newcommand{\Fref}[1]{Fig. \ref{#1}}

\newcommand{\Sref}[1]{Sec. \ref{#1}}

\newcommand{\Tref}[1]{Table \ref{#1}}

\newcommand{\R}{\mathbb{R}}

\usepackage[pdftex,breaklinks,colorlinks,
    citecolor=black, urlcolor=blue]{hyperref}

\title{Background Suppression Network \\for Weakly-supervised Temporal Action Localization}

\author{%
  Pilhyeon Lee\\
  Yonsei University\\
  lph1114@yonsei.ac.kr \\
  \And
  Youngjung Uh \\
  Clova AI Research, NAVER Corp. \\
  youngjung.uh@navercorp.com \\
  \And
  Hyeran Byun\thanks{Corresponding Author} \\
  Yonsei University\\
  hrbyun@yonsei.ac.kr \\
}

\begin{document}

\maketitle

\begin{abstract}
Weakly-supervised temporal action localization is a very challenging problem because frame-wise labels are not given in the training stage while the only hint is video-level labels: whether each video contains action frames of interest.
Previous methods aggregate frame-level class scores to produce video-level prediction and learn from video-level action labels. This formulation does not fully model the problem in that background frames are forced to be misclassified as action classes to predict video-level labels accurately.
In this paper, we design Background Suppression Network (BaS-Net) which introduces an auxiliary class for background and has a two-branch weight-sharing architecture with an asymmetrical training strategy. This enables BaS-Net to suppress activations from background frames to improve localization performance.
Extensive experiments demonstrate the effectiveness of BaS-Net and its superiority over the state-of-the-art methods on the most popular benchmarks – THUMOS'14 and ActivityNet. Our code and the trained model are available at \href{https://github.com/Pilhyeon/BaSNet-pytorch}{https://github.com/Pilhyeon/BaSNet-pytorch}.
\end{abstract}

\begin{figure*}[t]
  \centering
  \includegraphics[width=0.85\textwidth]{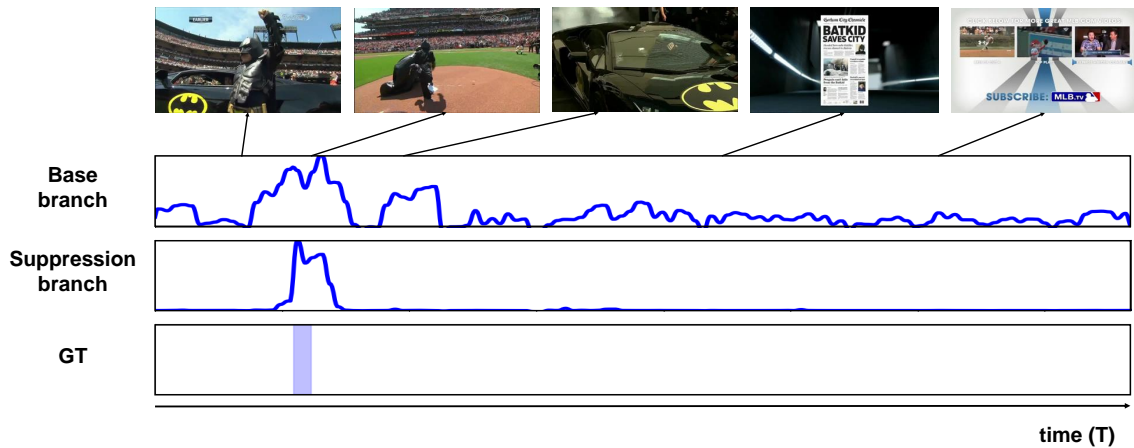}
  \caption{
  Visualization of the effectiveness of our method.
  Above is an example video from THUMOS'14 belonging to \textit{BaseballPitch} action. The first and second row are segment-wise activation sequences from Base branch and Suppression branch respectively, while the last row indicates ground truth (GT). 
  The horizontal axes denote timesteps of the video, while the vertical axes indicate the intensity of activation.
  }
  \label{fig:intro_figure}
\end{figure*}

\section{Introduction}
\label{sec:intro}

As the number of videos grows tremendously, extracting frames with actions from untrimmed videos is becoming more important so that humans can exploit them more efficiently. Furthermore, such frames are also useful data for machines to learn action representations. Accordingly, temporal action localization (TAL) has been developed to find frames containing actions in untrimmed videos, usually by training deep networks with full supervision, \ie, individual frames are labeled as action classes or background. However, training with full supervision has several pitfalls: they are (1) expensive (2) subjective (especially on action boundaries) (3) error-prone. Consequently, the research community has been interested in weakly-supervised temporal action localization (WTAL).

WTAL also aims to predict frame-wise labels but with weak supervision (\eg, video-level label, frequency of action instances in videos, or temporal ordering of action instances). Among them, the video-level label is the most commonly used weak supervision where each video is treated as a positive sample for action classes if it contains corresponding action frames. We note that a video can have multiple action classes as its label.
In order to disseminate the video-level label to individual frames, some previous methods formulate WTAL as multiple instance learning (MIL) which employs labels for bags of instances rather than those for individual instances~\cite{wang2017untrimmednets,paul2018w,Xu2019SegregatedTA}. As a video can be defined as a set of multiple frames, they first classify individual frames into action classes and then aggregate the frame-level scores to predict the video's action classes so that classification loss from video-level can guide frame-level predictions.

In this paper, we argue that previous MIL-based approaches do not fully model the problem in that background frames have not been regarded as a separate class although they do not belong to any action class. As a result, background frames are trained to be classified as action classes of the video to minimize loss from video-level even though they do not have certain features of actions. This inconsistency pushes background frames towards action classes, which causes false positives and performance degradation.

To tackle this problem, we introduce an auxiliary class for background frames. Since all untrimmed videos contain background frames, they are positive samples for their original action classes and the background class at the same time. The aforementioned inconsistency is resolved as all frames in a video now have their own categories to target.
We note that our approach is in line with fully-supervised methods for object detection~\cite{ren2015faster,redmon2016you,liu2016ssd} and TAL~\cite{shou2016temporal,zhao2017temporal} in employing the background class.
However, in weakly-supervised setting, introducing background class alone does not lead to improvement because we have no negative sample for background class to train. This means the network will eventually learn to produce high scores for background class regardless of input videos. 

Hence, to better exploit background class, we design Background Suppression Network (BaS-Net) containing two branches: Base branch and Suppression branch.
Base branch has the usual MIL architecture which takes frame-wise features as input and produces frame-wise class activation sequence (CAS) to classify videos as positive samples for their action classes and the background class.
Meanwhile, Suppression branch starts with a filtering module which is expected to attenuate input features from background frames, followed by the same architecture of Base branch with shared weights. Unlike Base branch, the objective of Suppression branch is to minimize scores for the background class for all videos while optimizing the original objective for the action classes. Because two branches share weights, they are restricted from optimizing both of their contrasting objectives at the same time given the same input. To resolve the restriction, the filtering module learns to suppress the activations from backgrounds. Finally, Suppression branch becomes free from the interference of background frames and, in result, localizes action more precisely.

The effectiveness of our method is illustrated in \Fref{fig:intro_figure}. Thanks to the filtering module, Suppression branch successes to suppress the activations from background frames and localize the action instance more accurately. In a later section, ablation study verifies that explicitly modeling background class and joint learning with the contrasting training objectives both are necessary to improve performance.

Our contributions are three-fold:

\begin{itemize}
    \item We introduce an auxiliary class representing background which was a missing element to model weakly-supervised temporal action localization problem.
    \item We propose an asymmetrical two-branch weight-sharing architecture with a filtering module and contrasting objectives to suppress activations from background frames.
    \item Our BaS-Net outperforms current state-of-the-art WTAL methods in experiments on the most popular benchmarks - THUMOS'14 and ActivityNet.
\end{itemize}

\begin{figure*}[t]
  \centering
  \includegraphics[clip=true, width=0.95\textwidth]{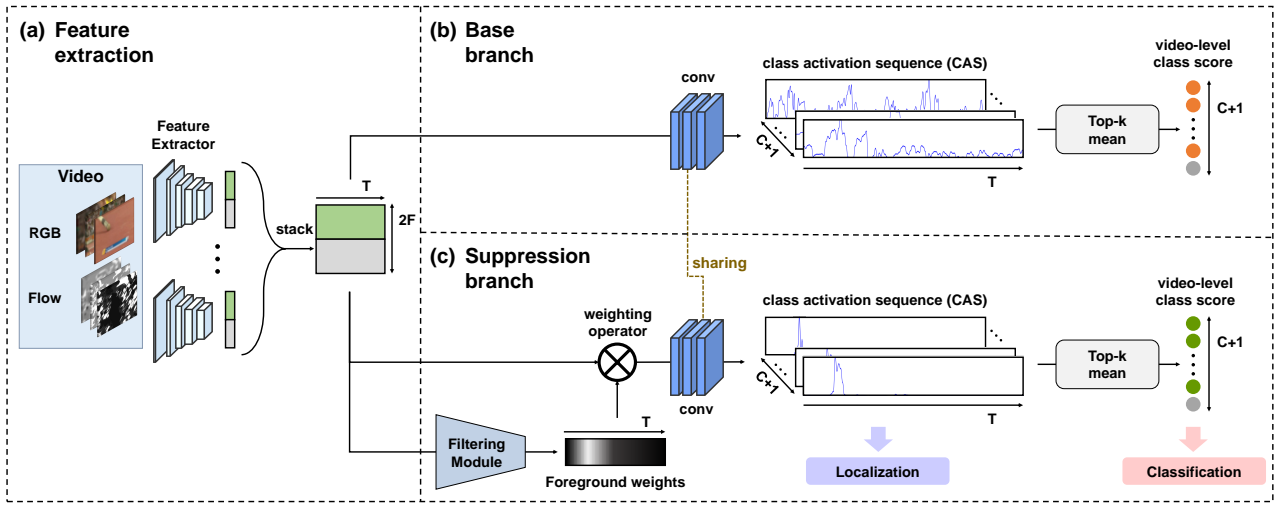}
  \caption{Overview of the proposed method. It consists of three parts: (a) Feature Extraction, (b) Base branch, and (c) Suppression branch}
  \label{fig:main_architecture}
\end{figure*}

\section{Related Work}

\paragraph{Fully-supervised temporal action localization (TAL)}

TAL is challenging because it requires not only action classes but also temporal intervals of the actions. To tackle the problem, previous methods mostly depend on full supervision, \ie, temporal annotations. Many of them~\cite{shou2016temporal,yuan2016temporal} generate proposals by sliding window and classify them into $C+1$ classes for $C$ action classes plus background class. Furthermore, several work~\cite{xu2017r,chao2018rethinking} attempts to generalize object detection algorithm to TAL. Most recently, a sophisticated proposal generation~\cite{lin2018bsn} and Gaussian temporal modeling~\cite{long2019gaussian} are proposed for accurate action localization.

\paragraph{Weakly-supervised temporal action localization (WTAL)}

WTAL solves the same problem but with less supervision, \eg, video-level labels. To derive frame-wise scores from video-level labels, previous methods generate class activation sequence (CAS). Some of them tackle the conventional problem that CAS tends to focus on a few discriminative frames~\cite{singh2017hide,Yuan2019MARGINALIZEDAA,liu2019completeness}.
While STPN~\cite{nguyen2018weakly} leverages class-agnostic attention weights along with CAS, Autoloc~\cite{shou2018autoloc} generates proposals by regression instead of thresholding.
Meanwhile, some work~\cite{wang2017untrimmednets,paul2018w,Xu2019SegregatedTA} formulates WTAL as multiple instance learning (MIL) problem
as we do.
However, as mentioned in \Sref{sec:intro}, they do not fully model WTAL problem in that they did not consider the background class so background frames are to be classified as any action class.
On the contrary, we introduce an auxiliary background class and also propose to suppress background frames for better action localization.

\section{Proposed Method}
\label{sec:proposed_method}

In this section, we describe details of our Background Suppression Network (BaS-Net). The overall architecture of BaS-Net is illustrated in \Fref{fig:main_architecture}. Before the detailed description, we first formulate weakly-supervised temporal action localization (WTAL) problem.

\paragraph{Problem Formulation}
Suppose that we are given $N$ training videos $\{v_{n}\}^{N}_{n=1}$ with their video-level labels $\{\mathbf{y}_{n}\}^{N}_{n=1}$, where $\mathbf{y}_{n}$ is $C$-dimensional binary vector with $y_{n;c}=1$ if $n$-th video contains $c$-th action category otherwise $0$ for $C$ classes. A video may contain multiple action classes, \ie, $\sum_{c=1}^{C}{y_{n;c}}\geq1$. Each input video goes through a network to generate frame-level class scores, \ie, class activation sequence (CAS). Afterwards, the scores are aggregated to produce a video-level class score. The network is trained to correctly predict video-level label, which is a proxy objective for CAS.
At test time, frame-wise action intervals are inferred by thresholding CAS for predicted action classes.

\subsection{Background class}
As discussed in \Sref{sec:intro}, without background class, activations from background frames lean towards action classes, which causes disturbance to accurate localization. In order to alleviate the disturbance, we introduce an auxiliary class representing the background. Then, naturally, all training videos are labeled as positive samples for background class since every untrimmed video contains background frames. This leads to a data imbalance problem where we have no negative sample for background class to use for training and corresponding CAS will always be high. Consequently, adding background class alone does not bring performance improvement, which is also verified by ablation study in \Sref{sec:experiments}.

\subsection{Two-branch Architecture}
Hence, we design a two-branch architecture to better exploit the background class. As illustrated in \Fref{fig:main_architecture}, our architecture contains two branches following a feature extractor; Base branch and Suppression branch. Both branches, sharing their weights, take a feature map and produce CAS to predict video-level scores with two differences:
\textit{i}) Suppression branch contains a filtering module which learns to filter out background frames to ultimately suppress activations from them in CAS.  \textit{ii}) Their training objectives are different. The objective of Base branch is to classify an input video as a positive sample for its original action classes and also for the background class. On the other hand, Suppression branch with the filtering module is trained to minimize the background class score with the same objective for original action classes.
The weight-sharing strategy prevents the branches from satisfying both of their objectives at the same time when the same input is given. Therefore, the filtering module is the only key to resolve the congested condition and is trained to suppress activations from background frames to pursue both objectives simultaneously. This reduces the interference of background frames and improves the action localization performance.

\subsection{Background Suppression Network}

\paragraph{Feature extraction}
We first divide each input video $v_{n}$ into 16-frame non-overlapping $L_{n}$ segments due to memory constraint, \ie, $v_{n}=\{{s}_{n,l}\}^{L_{n}}_{l=1}$. To deal with large variation of video lengths, we sample a fixed number of $T$ segments $\{\tilde{s}_{n, t}\}^{T}_{t=1}$ from each video.
Then, we feed sampled RGB and flow segments into the pre-trained feature extractor to generate $F$-dim feature vectors $x^{\text{RGB}}_{n,t} \in \R^{F}$ and $x^{\text{flow}}_{n,t} \in \R^{F}$, respectively. Afterwards, RGB and flow features are concatenated to build complete features $x_{n,t} \in \mathbb{R}^{2F}$, which are then stacked along temporal dimension to form a feature map of length $T$, \ie, $X_{n} = \begin{bmatrix}x_{n,1}, ..., x_{n,T}\end{bmatrix} \in \mathbb{R}^{2F\times T}$ (\Fref{fig:main_architecture} (a)).

\paragraph{Base branch}
To predict segment-level class scores, we generate CAS $\mathcal{A}_{n}$ where each segment has its class score by feeding the feature map into temporal 1D convolutional layers. This can be formalized as follows for a video $v_n$:
\begin{equation}
  \label{equ:cas_module}
  \mathcal{A}_{n} = f_{\text{conv}}(X_{n}; \phi)
\end{equation}
where $\phi$ denotes trainable parameters in the convolutional layers and $\mathcal{A}_{n}\in\mathbb{R}^{(C+1) \times T}$. $\mathcal{A}_{n}$ has $C+1$ dimensions because we use $C$ action classes and one auxiliary class for the background.

Afterwards, we aggregate segment-level class scores to derive a single video-level class score which will be compared to the ground truth. There are several approaches to gather scores and we adopt top-k mean technique following the previous work~\cite{wang2017untrimmednets,paul2018w}. Then, video-level class score for class $c$ of video $v_n$ can be derived as follows:
\begin{equation}
  \label{equ:top_k_mean}
  a_{n;c} = \text{aggregate}(\mathcal{A}_{n}, c) = \frac{1}{k} \max_{\substack{A\subset \mathcal{A}_{n}[c,:] \\ |A|=k}} \sum_{\forall a \in A}{a}
\end{equation}
where $ k = \floor[\Big]{\frac{T}{r}} $ and $r$ is a hyperparameter to control the ratio of selected segments in a video.

The video-level class score is then used to predict the probability of being a positive sample for each class by applying softmax function along class dimension:
\begin{equation}
  \label{equ:softmax}
  \mathbf{p}_{n} = \text{softmax}(\mathbf{a}_{n})
\end{equation}
where $\mathbf{p}_{n}$ has $C+1$ dimensions and each dimension indicates the probability of being a positive sample regarding its respective category for video $v_n$.

To train the network, we define a loss function $\mathcal{L}_{base}$ with binary cross-entropy loss for each class.
\begin{equation}
  \label{equ:loss_base}
  \mathcal{L}_{\text{base}} = \frac{1}{N}\sum_{n=1}^{N}\sum_{c=1}^{C+1}-y^{\text{base}}_{n;c}\log(p_{n;c})
\end{equation}
where $\mathbf{y}^{\text{base}}_{n} = \begin{bmatrix}y_{n;1}, ..., y_{n;C}, 1\end{bmatrix}^{T} \in \mathbb{R}^{C+1}$ is the video-level label for $n$-th video. The additional label for the background class is set to be positive considering that all training videos contain background frames.

\paragraph{Suppression branch}
Different from Base branch, Suppression branch contains a filtering module in its front, which is trained to suppress background frames by the opposite training objective for the background class. The filtering module consists of two temporal 1D convolutional layers followed by sigmoid function. The output of the filtering module is foreground weights $\mathcal{W}_{n}\in\R^{T}$ which range from 0 to 1. While the configuration of the filtering module is similar to the attention module in STPN~\cite{nguyen2018weakly}, it should be noted that training objectives are different so that its goal to target and what it learns are also different from STPN.
The foreground weights from the filtering module are multiplied to the feature map over the temporal dimension to filter out background frames. This step can be expressed as follows:
\begin{equation}
  \label{equ:filtering}
  {X'}_{n} = {X}_{n} \otimes \mathcal{W}_{n}
\end{equation}
where ${X'}_{n} \in \mathbb{R}^{2F \times T}$ and ${\otimes}$ denotes element-wise multiplication over temporal dimension.

The remaining process is analogous to Base branch except that the input feature map is different:
\begin{equation}
  \label{equ:filtered_cas}
  \mathcal{A'}_{n} = f_{\text{conv}}(X'_{n}; \phi)
\end{equation}
We note that the convolutional layers of two branches share weights. Following equations (\ref{equ:top_k_mean}) and (\ref{equ:softmax}), we obtain the video-level class score ${a'}_{n;c}=\text{aggregate}(\mathcal{A'}_{n}, c)$ and the class-wise probability $\mathbf{p'}_{n}=\text{softmax}(\mathbf{a'}_{n})$ where backgrounds are suppressed. 

We build the loss function $\mathcal{L}_{supp}$ with binary cross-entropy loss for each class.
\begin{equation}
  \label{equ:loss_supp}
  \mathcal{L}_{\text{supp}} = \frac{1}{N}\sum_{n=1}^{N}\sum_{c=1}^{C+1}-y^{\text{supp}}_{n;c}\log({p'}_{n;c})
\end{equation}
where $\mathbf{y}^{\text{supp}}_{n} = \begin{bmatrix}y_{n;1}, ..., y_{n;C}, 0\end{bmatrix}^{T} \in \mathbb{R}^{C+1}$.
we set the label for the background class to $0$, which is different from that of Base branch to train the filtering module to suppress background frames.

\paragraph{Joint training}
We jointly train Base branch and Suppression branch. The overall loss function we need to optimize is composed as follows:
\begin{equation}
  \label{equ:loss_total}
  \mathcal{L}_{\text{overall}} = \alpha \mathcal{L}_{\text{base}} + \beta \mathcal{L}_{\text{supp}} + \gamma \mathcal{L}_{\text{norm}}
\end{equation}
where $\alpha$, $\beta$, and $\gamma$ are the hyperparmaters. Following the previous work~\cite{nguyen2018weakly,Xu2019SegregatedTA}, we employ the L1 normalization of attention weights, \ie, $\mathcal{L}_{\text{norm}}=\frac{1}{N}\sum_{n=1}^{N}|\mathcal{W}_{n}|$, in order to make foreground weights more polarized.

\subsection{Classification and Localization}
After describing how our model is configured and trained, we turn to discuss how it works at test time. Since we suppress activations from background frames with our filtering module, it is reasonable to use the output of Suppression branch for inference. For the classification, we discard classes whose probabilities in ${p'_{n}}$ are below the threshold $\theta _{\text{class}}$. Then, for the remaining categories, we threshold the CAS with threshold $\theta _{\text{act}}$ to select candidate segments. Afterward, each set of consecutive candidate segments becomes a proposal. We compute the confidence score for each proposal using the contrast between inner and outer areas following the recent work~\cite{liu2019completeness}.

\begin{table*}[t]
\caption{
Comparison with the state-of-the-art methods on THUMOS'14. Entries are separated regarding the level of supervision. ${\dagger}$ indicates the use of additional labels, \ie, the number of action instances in videos. UNT and I3D denote the use of UntrimmedNets and I3D network as the feature extractor, respectively.
}
\begin{center}
\resizebox{.81\textwidth}{!}{
\begin{tabular}{c|l|ccccccccc}
\multirow{2}{*}{Supervision}       & \multicolumn{1}{c|}{\multirow{2}{*}{Method}} & \multicolumn{9}{c}{mAP@IoU}         \\
       & \multicolumn{1}{c|}{}                        & 0.1   & 0.2   & 0.3   & 0.4   & 0.5   & 0.6   & 0.7   & 0.8   & 0.9  \\ \hline\hline
\multirow{15}{*}{Full} & Richard et al.~\shortcite{richard2016temporal}   & 39.7  & 35.7  & 30.0  & 23.2  & 15.2  & -     & -     & -     & -    \\
       & S-CNN~\shortcite{shou2016temporal}      & 47.7  & 43.5  & 36.3  & 28.7  & 19.0  & 10.3  & 5.3  & -  & -  \\
       & Yeung et al.~\shortcite{yeung2016end}     & 48.9  & 44.0  & 36.0  & 36.0  & 36.0  & 26.4  & 17.1  & -  & -   \\
       & PSDF + T-SVM~\shortcite{yuan2016temporal}      & 51.4  & 42.6  & 33.6  & 26.1  & 18.8  & -     & -     & -     & -    \\
       & CDC~\shortcite{shou2017cdc}      & -  & -  & 40.1  & 29.4  & 23.3  & 13.1  & 7.9  & -  & -  \\
       & Yuan et al.~\shortcite{yuan2017temporal}      & 51.0  & 45.2  & 36.5  & 27.8  & 17.8  & -     & -     & -     & -    \\
       & CBR~\shortcite{gao2017cascaded}       & 60.1  & 56.7  & 50.1  & 41.3  & 31.0  & 19.1  & 9.9  & -  & -  \\
       & R-C3D~\shortcite{xu2017r}        & 54.5  & 51.5  & 44.8  & 35.6  & 28.9  & -     & -     & -     & -    \\
       & SSN~\shortcite{zhao2017temporal}      & 66.0  & 59.4  & 51.9  & 41.0  & 29.8  & -     & -     & -     & -    \\
       & SSAD~\shortcite{lin2017single}       & 50.1  & 47.8  & 43.0  & 35.0  & 24.6  & -     & -     & -     & -    \\
       & TPC~\shortcite{yang2018exploring}      & -  & -  & 44.1  & 37.1  & 28.2  & 20.6  & 12.7 & - & - \\
       & TAL-Net~\shortcite{chao2018rethinking}     & 59.8  & 57.1  & 53.2  & \textbf{48.5}                     & \textbf{42.8}                     & \textbf{33.8}                     & \textbf{20.8}                     & -                     & -                    \\
       & Action Search~\shortcite{alwassel2018action} & 51.8 & 42.4 & 30.8  & 20.2  & 11.1  & -     & -    \\
       & BSN~\shortcite{lin2018bsn}       & -  & -  & 53.5  & 45.0  & 36.9  & 28.4  & 20.0  & -  & - \\
       & GTAN~\shortcite{long2019gaussian}       & \textbf{69.1}  & \textbf{63.7}  & \textbf{57.8}                     & 47.2  & 38.8  & -  & -  & -  & - \\ \hline
\multirow{1}{*}{Weak${\dagger}$} & STAR~\shortcite{Xu2019SegregatedTA}        & \textbf{68.8}  & \textbf{60.0}  & \textbf{48.7}                     & \textbf{34.7}                     & \textbf{23.0}                     & -     & -  & -  & -    \\ \hline
\multirow{12}{*}{Weak} 
       & UntrimmedNet~\shortcite{wang2017untrimmednets}      & 44.4  & 37.7  & 28.2  & 21.1  & 13.7  & -     & -  & -  & -    \\
       & Hide-and-seek~\shortcite{singh2017hide}     & 36.4  & 27.8  & 19.5  & 12.7  & 6.8   & -     & -  & -  & -    \\
       & STPN (UNT)~\shortcite{nguyen2018weakly}    & 45.3  & 38.8  & 31.1  & 23.5  & 16.2  & 9.8   & 5.1   & 2.0   & 0.3    \\
       & AutoLoc~\shortcite{shou2018autoloc}      & -  & -  & 35.8  & 29.0  & 21.2  & 13.4  & 5.8  & -  & -  \\
       & W-TALC (UNT)~\shortcite{paul2018w}      & 49.0  & 42.8  & 32.0  & 26.0  & 18.8  & -     & 6.2  & -  & -  \\
       & Liu et al. (UNT)~\shortcite{liu2019completeness}      & 53.5  & 46.8  & 37.5  & 29.1  & 19.9  & 12.3  & 6.0  & -  & -  \\
       & Ours (UNT)            & \textbf{56.2}  & \textbf{50.3}  & \textbf{42.8} & \textbf{34.7} & \textbf{25.1} & \textbf{17.1} & \textbf{9.3}  & \textbf{3.7}  & \textbf{0.5} \\
       \cline{2-11}
       & STPN (I3D)~\shortcite{nguyen2018weakly}    & 52.0  & 44.7  & 35.5  & 25.8  & 16.9  & 9.9   & 4.3   & 1.2   & 0.1    \\
       & W-TALC (I3D)~\shortcite{paul2018w}      & 55.2  & 49.6  & 40.1  & 31.1  & 22.8  & -     & 7.6  & -  & -  \\
       & MAAN~\shortcite{Yuan2019MARGINALIZEDAA}      & \textbf{59.8}  & 50.8  & 41.1  & 30.6  & 20.3  & 12.0  & 6.9  & 2.6  & 0.2  \\
       & Liu et al. (I3D)~\shortcite{liu2019completeness}      & 57.4  & 50.8  & 41.2  & 32.1  & 23.1  & 15.0  & 7.0  & -  & -  \\
       & Ours (I3D)            & 58.2  & \textbf{52.3}  & \textbf{44.6} & \textbf{36.0} & \textbf{27.0} & \textbf{18.6} & \textbf{10.4}  & \textbf{3.9}  & \textbf{0.5}

\end{tabular}
}
\end{center}
\label{table:quant_thumos}
\end{table*}

\section{Experiments}
\label{sec:experiments}
In this section, we evaluate our BaS-Net with extensive experiments. We first describe details of the experimental settings, followed by comparison with the state-of-the-art methods and ablation study. Lastly, we visually demonstrate qualitative results of our method.

\subsection{Experimental Settings}
\paragraph{Dataset} We conduct experiments on weakly-supervised temporal action localization task on the most popular benchmarks: THUMOS’14~\cite{THUMOS14} and ActivityNet~\cite{caba2015activitynet}. They consist of untrimmed videos and provide both video-level action labels and frame-level temporal annotations. Note that we utilize only video-level labels for training and temporal annotations are used only for evaluation.

\paragraph{Evaluation Metrics} Following standard evaluation metrics, we measure mean average precision (mAP) at several different levels of intersection of union (IoU) thresholds. We employ the evaluation code provided by ActivityNet\footnote{https://github.com/activitynet/ActivityNet/} to evaluate methods on both datasets. 
 
\paragraph{Implementation Details}
We use two networks, namely UntrimmedNet~\cite{wang2017untrimmednets} and I3D networks~\cite{carreira2017quo}, as our feature extractor. They are pre-trained on ImageNet~\cite{deng2009imagenet} and Kinetics~\cite{carreira2017quo}, respectively. We note that the feature extractor is not fine-tuned for fair comparison. We use TVL1 algorithm~\cite{wedel2009improved} for generating optical flow of segments.

We fix the number of input segments $T$ to 750. To sample $T$ segments from each video, we use stratified random perturbation during training and uniform sampling during test, same as STPN~\cite{nguyen2018weakly}. All hyperparameters are empirically determined by grid search; $r=8$, $\alpha=1$, $\beta=1$, $\gamma=10^{-4}$, and $\theta _{\text{class}}=0.25$. For $\theta _{\text{act}}$, we use a set of thresholds from 0 to 0.5 with the step 0.025 and perform non-maximum suppression (NMS) with threshold 0.7 to remove highly overlapped proposals. Experiments are conducted on a single GTX 1080Ti GPU.

\begin{table*}[t]
\caption{
Effect of each component on the action localization performance on THUMOS'14. The column AVG denotes the average mAP under the IoU thresholds from 0.1 to 0.9
}
\begin{center}
\resizebox{.95\textwidth}{!}{
\begin{tabular}{c|ccc|cccccccccc}
                                                     & &                                                             &                                                            &  \multicolumn{10}{c}{mAP@IoU} \\
                                        \begin{tabular}[c]{@{}c@{}}\end{tabular} &
\begin{tabular}[c]{@{}c@{}}Base\\ branch\end{tabular} & \begin{tabular}[c]{@{}c@{}}background\\ class\end{tabular} & \begin{tabular}[c]{@{}c@{}}Suppression\\ branch\end{tabular} & 0.1 & 0.2 & 0.3             & 0.4             & 0.5            & 0.6            & 0.7 & 0.8 & 0.9 & AVG           \\ \hline\hline
baseline &  \checkmark                                                     &                                                            &                                                           & 32.3  & 25.2  & 19.8            & 15.9            & 12.0            & 8.7            & 4.7  & 1.4  & 0.2 & 13.4          \\
Base branch &  \checkmark                                                     & \checkmark                                                          &                                                             & 28.5  & 23.0  & 18.1            & 13.7            & 9.2            & 5.8            & 2.7  & 0.8  & 0.1 & 11.3           \\
Suppression branch  &                                                       & \checkmark                                                          & \checkmark                                                         & 49.1 & 42.5 & 33.5 & 26.0 & 18.6 & 12.8 & 6.2 & 2.0 & \textbf{0.5} & 21.2           \\
BaS-Net  &  \checkmark                                                     & \checkmark                                                          & \checkmark                                                         & \textbf{58.2} & \textbf{52.3} & \textbf{44.6} & \textbf{36.0} & \textbf{27.0} & \textbf{18.6} & \textbf{10.4} & \textbf{3.9} & \textbf{0.5} & \textbf{27.9}
\end{tabular}

}
\end{center}
\label{table:ablation_components}
\end{table*}

\begin{table}[t]
\caption{
Comparison on ActivityNet1.3 validation set. The entries with an asterisk are from ActivityNet Challenge while ${\dagger}$ denotes additional use of frequency of action instances in videos for training. The column AVG means the average mAP at IoU thresholds 0.5:0.05:0.95.
}
\begin{center}
\resizebox{.95\columnwidth}{!}{
\begin{tabular}{c|l|cccc}
\multicolumn{1}{c|}{\multirow{2}{*}{Supervision}} & \multicolumn{1}{c|}{\multirow{2}{*}{Method}} & \multicolumn{4}{c}{mAP@IoU} \\
\multicolumn{1}{c|}{}                             & \multicolumn{1}{c|}{}                        & 0.5     & 0.75     & 0.95     & AVG   \\ \hline\hline
\multirow{9}{*}{Full}                 & Singh et al.~\shortcite{singh2016untrimmed}*                                 & 34.5    & -        & -    & -      \\
                                                  & CDC~\shortcite{shou2017cdc}*                                  & 45.3    & 26.0    & 0.2    & 23.8   \\
                                                  & TCN~\shortcite{dai2017temporal}*                                   & 36.4    & 21.2    & 3.9    & -   \\
                                                  & Xiong et al.~\shortcite{xiong2017pursuit}*                                 & 39.1    & 23.5    & 5.5    & 24.0   \\
                                                  & SSAD~\shortcite{lin2017single}*                                   & 49.0    & 32.9    & 7.9    & 32.3   \\
                                                  & R-C3D~\shortcite{xu2017r}                                    & 26.8    & -        & -    & 12.7      \\
                                                  & TAL-Net~\shortcite{chao2018rethinking}                                  & 38.2    & 18.3    & 1.3    & 20.2   \\
                                                  & BSN~\shortcite{lin2018bsn}                                   & 52.5    & 33.5    & \textbf{8.9}    & 33.7   \\
                                                  & GTAN~\shortcite{long2019gaussian}                                   & \textbf{52.6}    & \textbf{34.1}    & \textbf{8.9}    & \textbf{34.3}   \\ \hline
\multirow{1}{*}{Weak${\dagger}$}                & STAR~\shortcite{Xu2019SegregatedTA}                                         & \textbf{31.1}    & \textbf{18.8}    & \textbf{4.7}    & -   \\ \hline
\multirow{4}{*}{Weak}                & STPN~\shortcite{nguyen2018weakly}                                         & 29.3    & 16.9    & 2.6    & -   \\
                                                  & MAAN~\shortcite{Yuan2019MARGINALIZEDAA}                                         & 33.7    & 21.9    & 5.5    & -   \\
                                                  & Liu et al.~\shortcite{liu2019completeness}                                         & 34.0    & 20.9    & \textbf{5.7}    & 21.2   \\
                                                  & Ours                                         & \textbf{34.5}    & \textbf{22.5}    & 4.9    & \textbf{22.2}  
\end{tabular}}
\end{center}
\label{table:quant_anet13}
\end{table}

\subsection{Comparison with state-of-the-art methods}
We compare our BaS-Net with current state-of-the-art fully-supervised and weakly-supervised approaches at the several IoU thresholds. The results on THUMOS'14, ActivityNet1.2 and 1.3 are summarized in \Tref{table:quant_thumos}, \Tref{table:quant_anet12} and \Tref{table:quant_anet13}, respectively. In the tables, methods at different levels of supervision are separated by horizontal lines for fair comparison. We note that STAR~\cite{Xu2019SegregatedTA} cannot be directly compared with our method\footnote{STAR is a weakly-supervised method yet its level of supervision is different from that of ours since they exploit additional annotations, \ie, frequency of action instances.}.

\Tref{table:quant_thumos} demonstrates the quantitative results on THUMOS’14 in chronological order. The lower two partitions are grouped by choice of the feature extractor: UntrimmedNet (UNT) and I3D.
Our method significantly outperforms all state-of-the-art methods at the same level of supervision, regardless of the feature extractor network.
We also compare our BaS-Net with fully-supervised approaches.
Even with a much lower level of supervision, our method shows the least gap regarding the latest fully-supervised methods. Furthermore, it can be noticed that our method even outperforms several fully-supervised methods at some IoU thresholds.

We also evaluate our BaS-Net on ActivityNet1.3 in \Tref{table:quant_anet13}. We see that our method outperforms all other weakly-supervised approaches. Moreover, despite using weaker labels, our algorithm outperforms STAR at all IoU thresholds.

Experimental results on ActivityNet1.2 are shown in \Tref{table:quant_anet12} to compare our method with more methods. Our model outperforms all weakly-supervised methods, following the fully-supervised method with a small gap.

\begin{table}[t]
\caption{
Comparison with other methods on ActivityNet1.2 validation set. The column AVG shows the average mAP at IoU thresholds 0.5:0.05:0.95.
}
\begin{center}
\resizebox{.95\columnwidth}{!}{
\begin{tabular}{c|l|cccc}
\multicolumn{1}{c|}{\multirow{2}{*}{Supervision}} & \multicolumn{1}{c|}{\multirow{2}{*}{Method}} & \multicolumn{4}{c}{mAP@IoU} \\
\multicolumn{1}{c|}{}                             & \multicolumn{1}{c|}{}                        & 0.5     & 0.75     & 0.95     & AVG   \\ \hline\hline
\multirow{1}{*}{Full}                 & SSN~\shortcite{zhao2017temporal}                                   & \textbf{41.3}    & \textbf{27.0}    & \textbf{6.1}    & \textbf{26.6}   \\ \hline
\multirow{4}{*}{Weak}                
                                                  & AutoLoc~\shortcite{shou2018autoloc}                                         & 27.3    & 15.1    & 3.3    & 16.0   \\
                                                  & W-TALC~\shortcite{paul2018w}                                         & 37.0    & -    & -    & 18.0   \\
                                                  & Liu et al.~\shortcite{liu2019completeness}                                         & 36.8    & 22.0    & \textbf{5.6}    & 22.4   \\
                                                  & Ours                                         & \textbf{38.5}    & \textbf{24.2}    & \textbf{5.6}    & \textbf{24.3}  
\end{tabular}}
\end{center}
\label{table:quant_anet12}
\end{table}

\begin{table}[t]
\caption{
Performances for detecting background frames on THUMOS'14 (F-measure).
}
\begin{center}
\resizebox{.95\columnwidth}{!}{
\begin{tabular}{c|c|c|c}
\multicolumn{1}{c|}{}   & Base branch        & Suppression branch     & BaS-Net   \\ \hline\hline
\multicolumn{1}{c|}{F-measure}   & 0.541    & 0.775    & \textbf{0.846}

\end{tabular}}
\end{center}
\label{table:ablation_fmeasure}
\end{table}

\begin{figure*}[t]
  \centering
  \includegraphics[clip=true, width=0.9\textwidth]{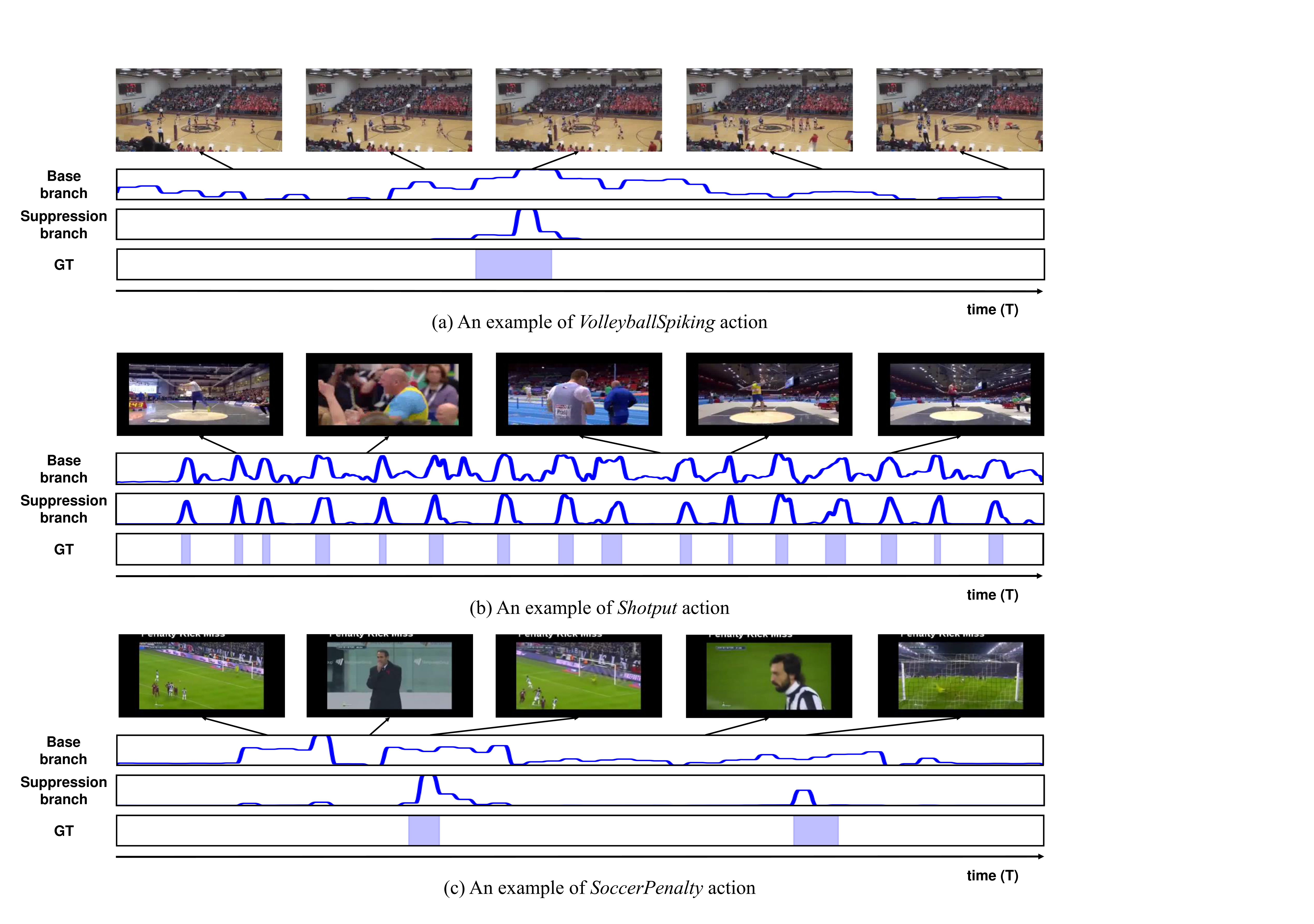}
  \caption{
  Qualitative results on THUMOS'14. For each example, there are three plots with several sampled frames.
  The first and second plot represent segment-wise activation sequences of the corresponding action from Base branch and Suppression branch, respectively.
  The last plot indicates the ground truths.
  The horizontal axes in the plots is the timesteps of the videos, while the vertical axes in the first two plots indicate the activation intensity which ranges from 0 to 1.
  }
  \label{fig:qualtitative_figure}
\end{figure*}

\subsection{Ablation study}
In \Tref{table:ablation_components}, We conduct ablation study on THUMOS’14 to investigate the contributions of different components of BaS-Net.

\begin{itemize}
    \item \textbf{Baseline.} We set the baseline with vanilla MIL setting, \ie, Base branch without the auxiliary background class.
    \item \textbf{Base branch.} We add an auxiliary class for background into the baseline, \ie, Base branch. As shown, the performance does not improve, rather decreases. We conjecture that it is because the network is trained to always produce high activations of the background class for any video due to the lack of negative samples, causing disturbance to classification. It indicates that solely correcting WTAL problem setting by explicitly modeling the background class cannot lead to performance improvement.
    \item \textbf{Suppression branch.} We evaluate a variant with only Suppression branch in order to assess the role of Base branch. With the filtering module acting like attention, it improves the localization performance from the baseline. However, we note that it is not derived from the background modeling, since there is no positive sample for background class.
    \item \textbf{BaS-Net.} By employing both branches and jointly training them with contrasting objectives, BaS-Net learns the background class as well as action classes and shows the best performance with large gaps from the others.
\end{itemize}

We also perform experiments on how effective each branch is for detecting background frames by measuring F-measure. \Tref{table:ablation_fmeasure} demonstrates that BaS-Net requires joint learning of both branches. 

\subsection{Qualitative results}
\Fref{fig:qualtitative_figure} shows several qualitative results on THUMOS'14.

\begin{itemize}
    \item \textbf{Sparse case.} \Fref{fig:qualtitative_figure} (a) is a challenging example because humans look small and actions sparsely occur \ie, background frames occupy a large portion of the video. Despite these challenges, our method successfully suppresses the activation from background frames and further seeks the action interval precisely.
    \item \textbf{Frequent case.} In \Fref{fig:qualtitative_figure} (b), the video has significantly frequent actions of \textit{Shotput}, which makes the localization difficult. Nonetheless, by distinguishing actions from the background, our method can accurately find the actions.
    \item \textbf{Challenging background case.} \Fref{fig:qualtitative_figure} (c) shows an example with challenging background which has a very similar appearance to foreground. As a result, in Base branch, some background frames show even higher activation than foreground frames. Even so, our Suppression branch successfully attenuate background activations, indicating that explicitly modeling background is important.
\end{itemize}

\section{Conclusion}

In this work, we identified a problem posed by the lack of background modeling in the previous weakly-supervised temporal action localization methods. To solve the problem, we proposed to classify not only action classes but also the background class in multiple instance learning.
Moreover,
to better exploit background information, we introduced a new two-branch architecture and asymmetrical training strategy.
Ablation study showed that the background class and the training strategy both are necessary to achieve performance improvement.
Through the extensive experiments, we demonstrated that our framework is effective for suppressing background and outperforms the current state-of-the-art methods for weakly-supervised temporal action localization task on both THUMOS’14 and ActivityNet.

\section{Acknowledgement}
{\small
This project was partly supported by Next-Generation Information Computing Development Program through the National Research Foundation of Korea (NRF) funded by the Ministry of Science and ICT (NRF-2017M3C4A7069370) and the Institute for Information \& Communications Technology Planning \& Evaluation (IITP) grant funded by the Korea government (No. 2019-0-01558: Study on audio, video, 3d map and activation map generation system using deep generative model)
}

{
\fontsize{9pt}{10pt} \selectfont
\bibliography{BaS-Net_arxiv}
\bibliographystyle{aaai}
}

\end{document}